\documentclass[letterpaper,10pt,conference]{ieeeconf}

\IEEEoverridecommandlockouts
\usepackage{xparse} 
\usepackage{amsmath} 
\usepackage{amssymb}
\usepackage{amsfonts} 
\usepackage{color}
\usepackage{siunitx}
\usepackage{calc}

\usepackage{graphicx}
\usepackage{subcaption}

\usepackage{hyperref}
\usepackage{cleveref}

\newcommand{\mbf}[1]{\mathbf{#1}}

\DeclareMathAlphabet{\mbfh}{OML}{cmm}{b}{it}

\newcommand{\norm}[1]{\left\Vert#1\right\Vert}

\newcommand{\bbm}{\begin{bmatrix}}
\newcommand{\ebm}{\end{bmatrix}}

\title{\LARGE \bf The Phoenix Drone: An Open-Source Dual-Rotor\\ Tail-Sitter Platform for Research and Education}

\author{Yilun Wu$^1$, Xintong Du$^2$, Rikky Duivenvoorden$^1$, and Jonathan Kelly$^1$
	\thanks{$^{1}$Yilun Wu, Rikky Duivenvoorden, and Jonathan Kelly are with the Space Terrestrial Autonomous Robotic Systems (STARS) Laboratory at the University of Toronto Institute for Aerospace Studies (UTIAS), Canada \texttt{\{yl.wu, rikky.duivenvoorden\}@robotics.utias.utoronto.ca, jkelly@utias.utoronto.ca}.}
	\thanks{$^{2}$Xintong Du is with the Dynamic Systems Laboratory (DSL) at the University of Toronto Institute for Aerospace Studies (UTIAS), Canada \texttt{xintong.du@robotics.utias.utoronto.ca}.}}

\begin{document}
\maketitle 
\thispagestyle{empty}
\pagestyle{empty}

\begin{abstract}
In this paper, we introduce the Phoenix drone: the first completely open-source tail-sitter micro aerial vehicle (MAV) platform. The vehicle has a highly versatile, dual-rotor design and is  engineered to be low-cost and easily extensible/modifiable. Our open-source release includes all of the design documents, software resources, and simulation tools needed to build and fly a high-performance tail-sitter for research and educational purposes.

The drone has been developed for precision flight with a high degree of control authority. Our design methodology included extensive testing and characterization of the aerodynamic properties of the vehicle. The platform incorporates many off-the-shelf components and 3D-printed parts, in order to keep the cost down. Nonetheless, the paper includes results from flight trials which demonstrate that the vehicle is capable of very stable hovering and accurate trajectory tracking.

Our hope is that the open-source Phoenix reference design will be useful to both researchers and educators. In particular, the details in this paper and the available open-source materials should enable learners to gain an understanding of aerodynamics, flight control, state estimation, software design, and simulation, while experimenting with a unique aerial robot.
\end{abstract}

\section{Introduction}

With the ever-increasing performance and decreasing cost of flight electronics, sensors and batteries, micro aerial vehicles (MAVs) are now deployed in a wide variety of domains, from agriculture to search and rescue. Vertical Take-Off and Landing (VTOL) platforms are particularly attractive for many applications because they combine the agility and maneuverability of rotary-wing vehicles with the efficiency and endurance of fixed-wing aircraft \cite{drone-review}. The tail-sitter, a type of VTOL platform that uses only two propellers and two actuated control surfaces (operated under the downwash of the propellers), is typically favoured over tilt-rotor and tilt-wing designs for its mechanical simplicity. Successful commercial implementations include the Wingtra mapping and surveying drone \cite{wingtra} and the Google X Project Wing delivery drone \cite{project-wing}, for example.

\begin{figure}[t!]
\vspace{2mm}
\centering
\setlength{\fboxsep}{0pt}%
\setlength{\fboxrule}{1pt}%
\fbox{\includegraphics[width=\columnwidth - 2pt]{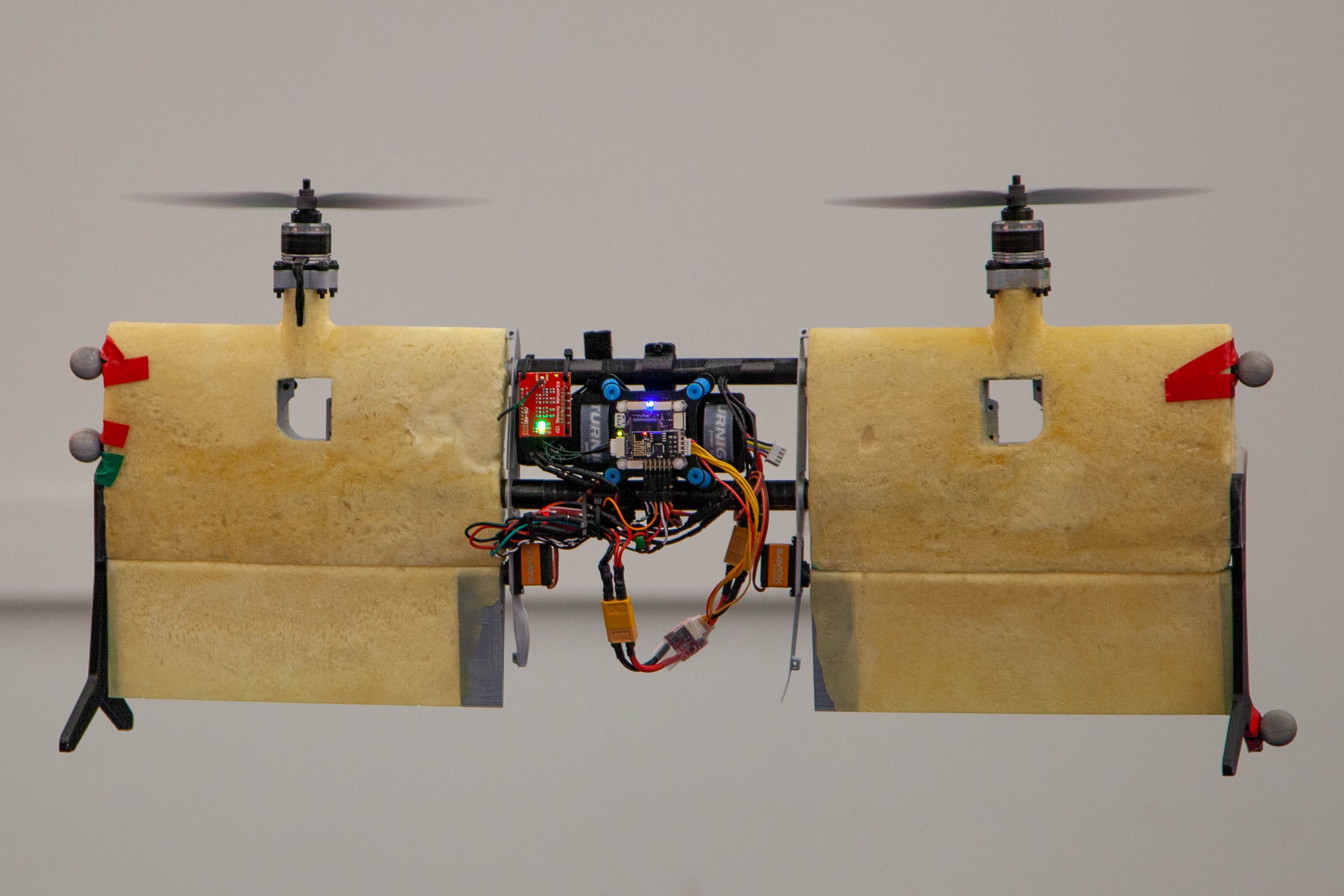}}
    \caption{The Phoenix drone, our open-source dual-rotor tail-sitter vehicle, in hovering flight in our laboratory.}
    \label{fig:HB}
\vspace{-\baselineskip}
\end{figure}

The availability of open-source MAV software libraries and hardware designs (e.g., the PX4 autopilot \cite{px4}) have enabled researchers, educators, and hobbyists to quickly prototype and test aerial robots without the burden of starting from scratch. This has led to a thriving MAV development community that continues to grow. However, to date, the vast majority of the accessible reference designs and associated tools have been for quadrotor vehicles, which are relatively easy to build and fly. In contrast, there are very limited resources available for individuals who wish to assemble and test VTOL platforms.
 
 To fill the above gap, in this paper we introduce the Phoenix drone, shown in \autoref{fig:HB}, which is, to the best of our knowledge, the first fully open-source tail-sitter vehicle for research and educational use. Our open-source package, available on GitHub\footnote{\url{https://github.com/utiasSTARS/PhoenixDrone}}, includes mechanical design documents, component lists, a carefully tuned and verified dynamics model, control software, and a full set of simulation tools---in short, everything necessary to understand, construct, test and verify a prototypical tail-sitter MAV.

Our goal is to enable researchers and educators to build high-performance tail-sitter vehicles easily, by leveraging online resources and taking advantage of modern digital manufacturing techniques. Towards this end, the Phoenix drone utilizes off-the-shelf actuators and computing hardware. The vehicle frame design incorporates a cast polyurethane foam core and 3D-printed plastic parts. We adopt the widely-used PX4 middleware to support our custom flight control software. We also include a MATLAB Simulink model and software-in-the-loop (SITL) Gazebo plugins to seamlessly compile and test flight code on the desktop.

The remainder of the paper is organized as follows. We summarize relevant, existing literature and related community resources for other types of platforms in Section \ref{sec:related-work}. We then describe the system design of our MAV and discuss our methodology in Section \ref{sec:design}. Section \ref{subsec:mech-design} focusses on mechanical design, with an emphasis on vehicle sizing and airfoil selection, while Section \ref{subsec:dynamic} introduces our dynamic modelling results, and Section \ref{sec:control-arch} gives an overview of the on-board control architecture. In Section \ref{sec:performance}, we present a series of flight experiments to characterize and benchmark the performance of the drone in the near-hover regime. Finally, we provide brief descriptions of the resources available in our open-source package in Section \ref{sec:resources} and consider several use cases in Section \ref{sec:conclusion}.

\section{Related Work}
\label{sec:related-work}

The tail-sitter is a novel MAV configuration with a substantial corpus of existing literature on vehicle design. Stone et al.\ conducted the earliest dual-rotor tail-sitter flight tests \cite{T-Wing}, achieving a hovering accuracy of roughly 1 m under a 6-8 knot winds. Bapst et al.\ \cite{Roman} developed the earliest dual-rotor tail-sitter MAV, derived a first-principles model of vehicle dynamics, and proposed a cascaded control architecture (now commonly used in MAVs). Later, Verling et al.\ \cite{Full-Att-TS} devised a modified tail-sitter with a customized airframe (modelled based on wind tunnel measurements); this design is closed due to its commercial association with Wingtra \cite{wingtra}. Ritz et al.\ \cite{Robin-Global} also developed a customized tail-sitter design, implementing optimal control at the attitude level to enable recovery from large attitude errors. Various control and estimation techniques have been applied to these vehicles to improve hovering performance when dealing with wind disturbances \cite{wind,disturb-reject}. Chiappinelli and Nahon \cite{ts-modelling} have also developed a basic modelling and control framework for tail-sitter vehicles. Alternative designs, such as quadrotor tail-sitters \cite{quad-tailsitter,quad-vtol} have been explored as well.

The widely-used PX4 \cite{px4} autopilot software is capable of controlling a dual-rotor VTOL tail-sitter, where the internal model employed is based on a modified TBS Caipirinha flying wing. We found during our early experiments, however, that the relatively small elevon size of the Caipirinha vehicle prevents it from performing aggressive or precise flight maneuvers. The use of a generic multi-copter attitude controller which maps pitch and yaw commands proportionally to elevon deflections ignores the coupling between propeller thrust/reaction torque and differential elevon deflections. Although the system can be stabilized in certain situations with carefully tuned gains, the controller is not ideal for precision flight.

The open-source flight stack and middleware which comes with PX4 has proven to be a very popular tool for enabling MAV research and for educational purposes. We note that other open-source robotics packages, such as the Duckietown \cite{Duckietown} autonomous vehicle testbed, have been very successful in the education space. Resources to build the Crazyflie \cite{crazyflie} nano-quadcopter are available in open-source form, and the recently-released PiDrone package \cite{PiDrone-IROS} also provides an open-source implementation of an easy-to-build quadrotor capable of indoor flight. The success of these examples clearly shows that open-source reference designs, available for free, are highly valuable to the robotics community (and thus that further releases should be encouraged).

Although the literature contains multiple examples describing the design and control of dual-rotor tail-sitter vehicles, to date, none of the designs have been released in open-source form. We reiterate that our goal is to provide a description of the design and development of a versatile tail-sitter and also to give researchers and educators a complete set of resources to simulate, build, and fly such a platform.

\section{System Design}
\label{sec:design}

The Phoenix drone is based upon the PX4 autopilot and uses the open-source PixRacer flight computer and PX4 middleware to support both flight control and SITL simulation. The PixRacer flight computer incorporates a 168 MHz ARM\textsuperscript{\textregistered} Cortex M4F microprocessor, which executes all control loops in real time. We use MAVROS, an open-source Robot Operating System (ROS) package, to communicate with the vehicle from our ground station (a laptop or desktop) over the MAVLink communication protocol. 

On board our prototype vehicle, two DYS-SN20A ESCs (electronic speed controllers), running modified BLHeli firmware, drive the twin TMotor 2208-18 1100 Kv brushless DC motors. Each motor is attached to a Gemfan 8-inch diameter 4.5-inch pitch propeller, capable of generating a maximum of 500 g of thrust; this leaves sufficient overhead for aggressive control of the vehicle, which has a mass of 650 g in total. Using its standard 3S 2200 mAh Li-Po battery, the drone has an endurance of approximately 20 minutes while hovering.

\subsection{Mechanical Design}
\label{subsec:mech-design}

As mentioned in Section \ref{sec:related-work}, we found that the non-symmetrical airfoil shape, relatively weak motor mounts, and limited elevon rotational range all made the TBS airframe unsuitable for high-performance indoor flight. To improve flight capabilities, we decided instead to design our own airframe. The Phoenix features a E168 low-Reynolds-number symmetrical airfoil with a span of 21 cm on each side. The full span of the wing was adjusted to accommodate the 8-in propellers installed over each wing surface, such that all of the control surface area is subject to propeller downwash. We chose a symmetrical airfoil to allow the vehicle to hover at zero pitch angle (i.e., with the propellers pointed exactly vertically). The low-Reynolds-number airfoil was adopted to minimize the effects of separation bubbles, which introduce nonlinearity in the aerodynamics around equilibrium when subject to low-speed propeller downwash.

\begin{figure}[t!]
\centering
\setlength{\fboxsep}{0pt}%
\setlength{\fboxrule}{1pt}%
\fbox{\includegraphics[width=\columnwidth - 2pt]{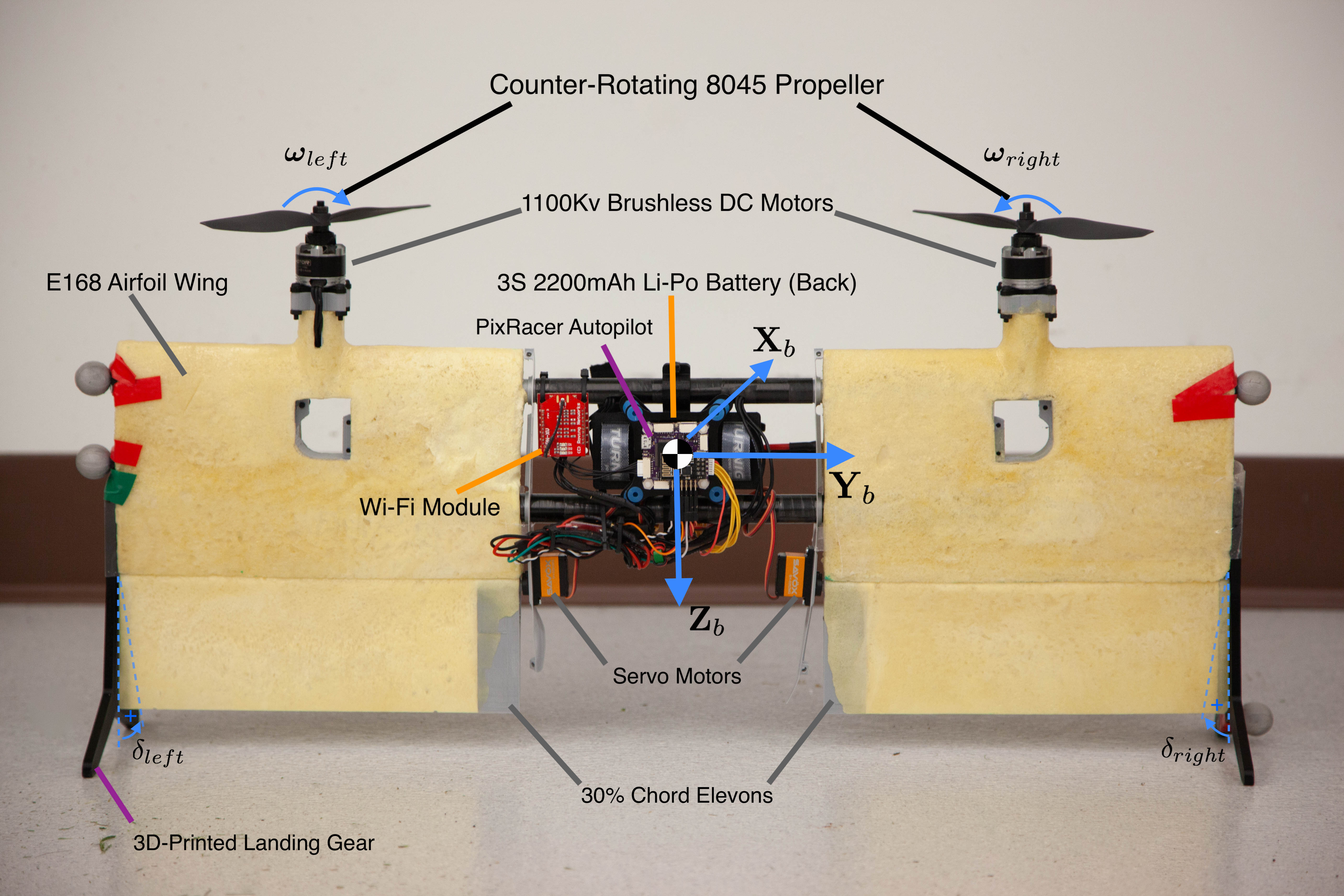}}
    \caption{Labelled vehicle components and body frame axis convention.}
    \label{fig:dyn-convention}
\vspace{-\baselineskip}
\end{figure}

To save weight while keeping the vehicle robust to shattering during crashes, the wings and elevons are cast in 2 pcf.\ polyurethane foam, leading to a final density of 0.055 g/cm$^3$. The two parts of the wing are joined together by two lightweight carbon fibre tubes (see \autoref{fig:dyn-convention}). All of the other parts, including the landing gear and the electronics housing, are 3D-printed. The total mass of the mechanical airframe by itself is 200 g.

\subsection{Dynamic Characterization}
\label{subsec:dynamic}

In this section, we introduce our dynamic model characterization of the Phoenix vehicle in its hovering regime. All vectors and matrices are expressed in the body frame, with axes defined according to \autoref{fig:dyn-convention} unless otherwise noted. We model the vehicle dynamics in terms of the forces and torques applied by the motors and the control surfaces: the forces and torques from the propellers are denoted by $\mbf{f}_{prop,i}$ and $\mbf{M}_{prop,i}$, respectively; forces and torques induced by airflow over the wings and control surfaces are denoted by $\mbf{f}_{aero,i}$ and $\mbf{M}_{aero,i}$, respectively. The subscript $i \in \{1,2\}$ refers to the left or right side of the vehicle, respectively. The overall forces and torques are
\begin{align}
	\label{eq:general-dyn}
	\mbf{f}_{tot} & = \sum_{i}(\mbf{f}_{prop,i} + \mbf{f}_{aero,i}) + \mbf{R}_w^bm\mbf{g}, \\
	\mbf{M}_{tot} & = \sum_{i}(\mbf{M}_{prop,i} + \mbf{M}_{aero,i}),
\end{align}
where
\begin{align}
\label{eq:dynamics}
\renewcommand{\arraystretch}{1.25}
	\mbf{f}_{prop,i} & = \bbm 0\\ 0\\ -k_t\boldsymbol{\omega}_{prop,i}^2 \ebm,\;\;\,
	\mbf{M}_{prop,i} = \bbm 0 \\ 0\\ \pm k_m\boldsymbol{\omega}_{prop,i}^2 \ebm, \\[2mm]
	\mbf{f}_{aero,i} & = \bbm -k_l\boldsymbol{\omega}_{prop,i}^2\delta_i \\ 0 \\k_d\boldsymbol{\omega}_{prop,i}^2\delta_i^2\ebm,\;
	\mbf{M}_{aero,i} =  \bbm 0 \\ -k_p\boldsymbol{\omega}_{prop,i}^2\delta_i  \\ 0\ebm.
\end{align}

As in most quadrotor systems, we model the propeller thrust and torque as a quadratic function of the rotational speed. We capture the lift, drag and pitch moments generated by the wing and control surfaces within the aerodynamic model; the forces and torques applied about other axes are negligible.

It is worth noting that the aerodynamic model proposed here is significantly simpler than those introduced in \cite{Robin-Global} and \cite{Full-Att-TS}, due to the usage of a symmetric airfoil (which eliminates non-zero bias). For a generic first-principles derivation of the forces/moments on such vehicles, see \cite{Roman}.

\begin{figure}
  \centering
  \vspace{-3mm}
  \begin{subfigure}[]{\linewidth}
      \includegraphics[width=\linewidth]{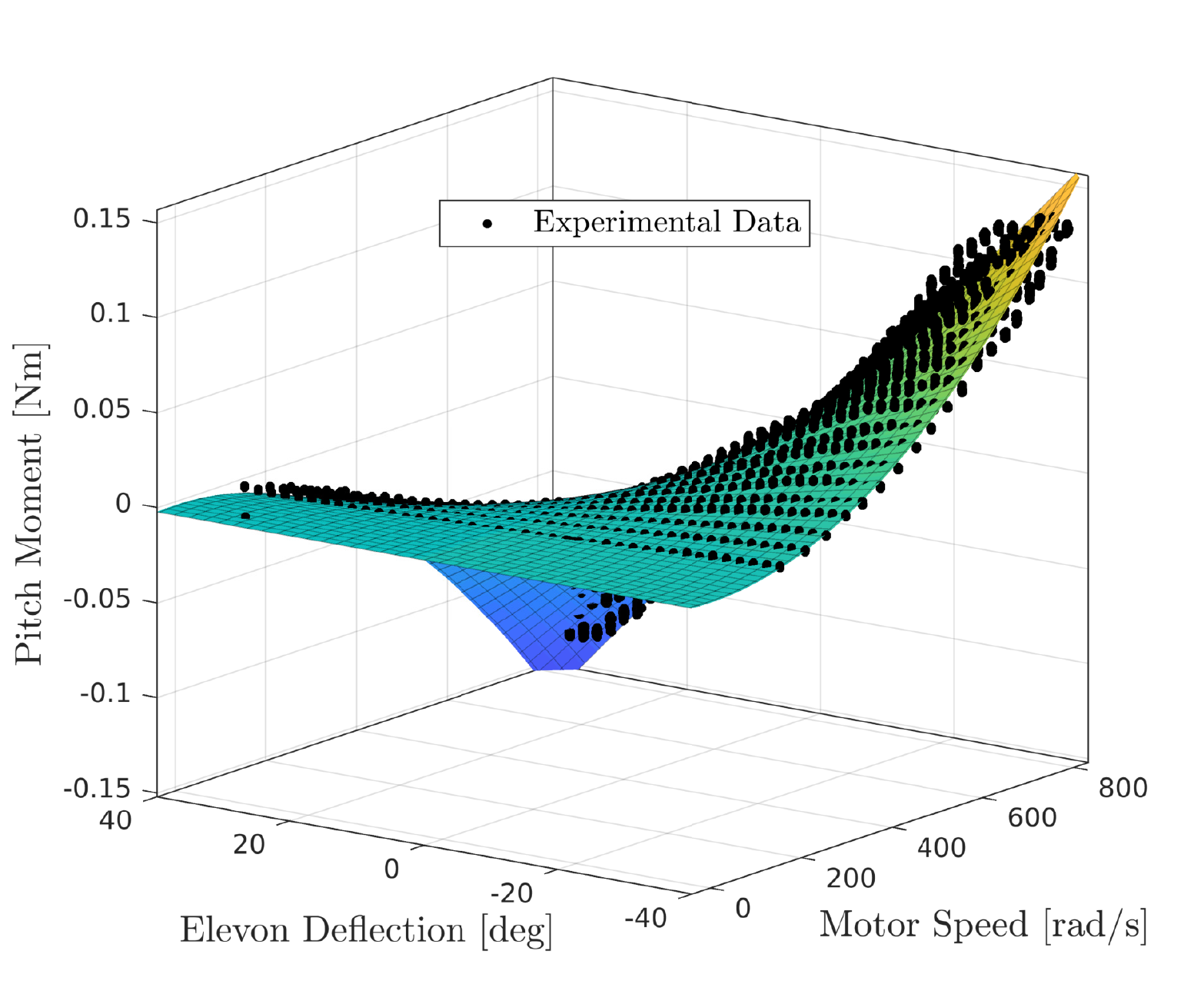}
    \caption{Pitch torque vs propeller speed and elevon deflection.}
  \end{subfigure}
  \vspace{2mm}
  \begin{subfigure}[]{0.47\linewidth}
    \includegraphics[width=\linewidth]{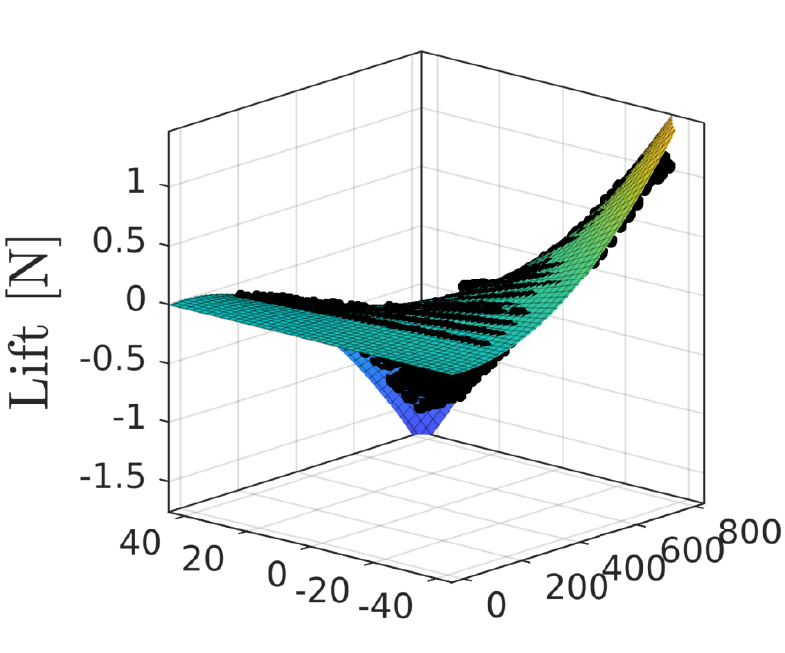}
     \caption{Lift vs propeller speed and elevon deflection.}
  \end{subfigure}
  \hfill
  \begin{subfigure}[]{0.47\linewidth}
    \includegraphics[width=\linewidth]{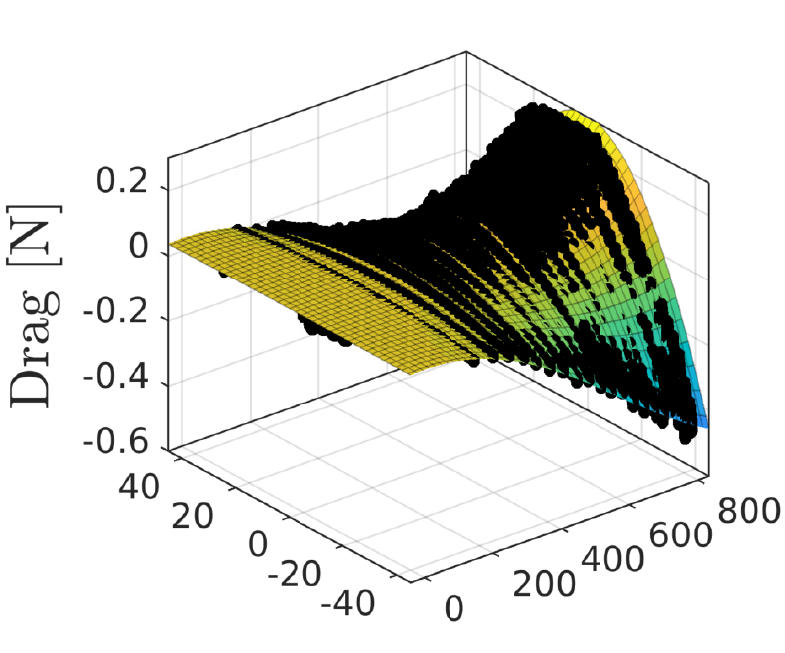}
    \caption{Drag vs propeller speed and elevon deflection.}
  \end{subfigure}
  \caption{Wing aerodynamic identification measurements and corresponding fitted models.}
  \label{fig:TS-Aero}
\vspace{-\baselineskip}
\end{figure}

The dynamic model parameters of the vehicle were established through extensive static tests using a 6-DOF force/torque load cell. Based on the force and torque measurements at varying propeller speeds and control surface deflections, parameters were extracted that gave a best fit to the experimental data. Raw measurements and the fitted model are shown in \Cref{fig:TS-Aero}. For the corresponding parameter values, please refer to \autoref{table:params}.

\subsection{Control Architecture}
\label{sec:control-arch}

\begin{figure}[h!]
	\includegraphics[width=1\linewidth]{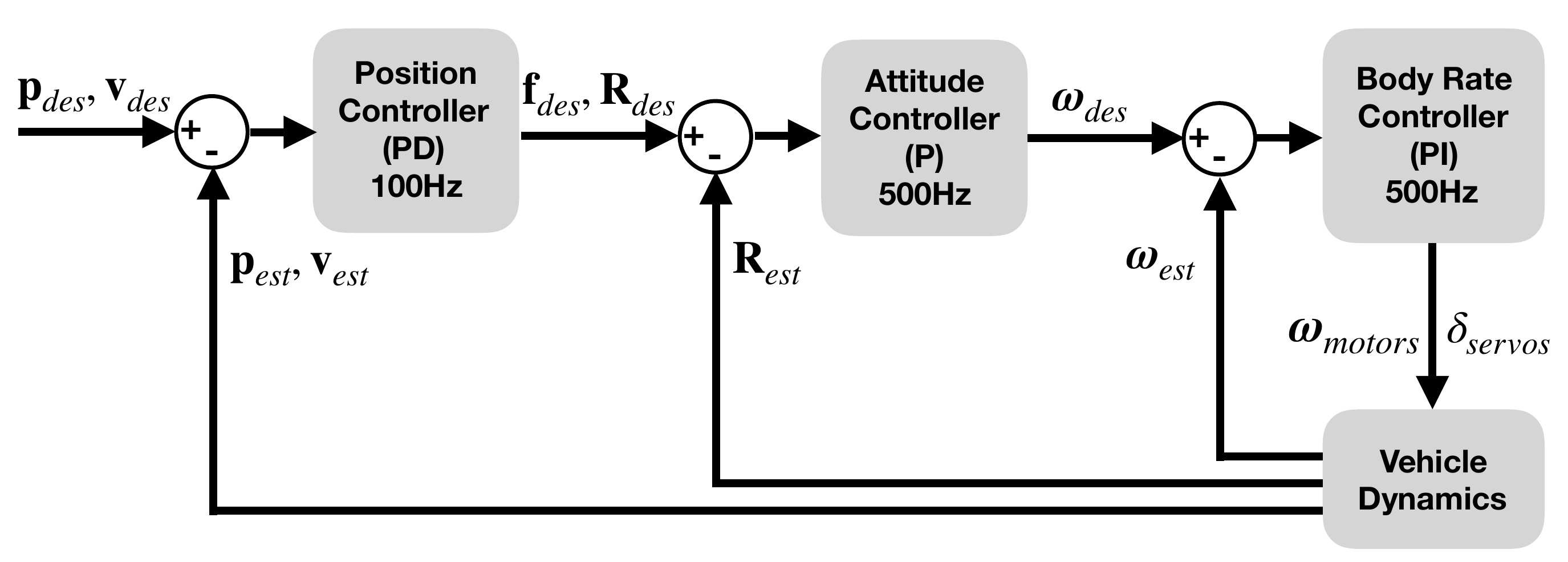}
	\caption{Complete motion control architecture, illustrating the three-controller hierarchy.}
	\label{fig:control-arch}
\vspace{-\baselineskip}
\end{figure}

The motion control architecture for the vehicle utilizes the cascaded control strategy shown in  \autoref{fig:control-arch}. This architecture includes three separate controllers running at different levels and frequencies: a high-level trajectory/position controller, a mid-level attitude controller, and a low-level rate controller. The controllers assume that information about the vehicle position, velocity, attitude, and attitude rates is available at a relatively high frequency. This data is provided by the native PX4 state estimator, which fuses onboard IMU readings with external pose measurements (from, e.g., a motion capture system or an on-board camera).

\subsubsection{Position Control}

The desired acceleration vector $\mbf{a}_{des}$ is determined based on the current position and velocity error. This outer loop runs at 100 Hz, synchronized to the Kalman filter that provides the position state estimates. Through the control law, the position dynamics behave as a second-order system with a desired time constant $\tau_p$ and a damping ratio $\zeta_p$. The desired acceleration and force vectors are given by
\begin{align}
	\mbf{a}_{des} &= -\mbf{g} + \frac{1}{\tau_p^2}(\mbf{p}_{des} - \mbf{p}_{est}) + \frac{2\zeta_p}{\tau_p}(\mbf{v}_{des} - \mbf{v}_{est}), \\
	\mbf{f}_{des} &= m\,\mbf{a}_{des},
\end{align}
\noindent where all vectors above are expressed in the inertial frame.

\subsubsection{Attitude Control}

The attitude controller attempts to align the vehicle heading to the desired target heading. Here, we assume that the vehicle is operating in the hover regime and that the reference heading $\psi$ should be maintained at all times. Heading alignment results in a z-axis rotation $\mbf{R_z}$; this is followed by a second rotation $\mbf{R_{xy}}$ about the body $x$ and $y$ axes to align the thrust direction as required. The two rotations are applied in sequence to determine the desired (full) rotation matrix $\mbf{R}_{des}$; individual rotation matrices and the necessary thrusts for these steps are obtained as follows:
\begin{align}
	\mbf{R}_z &= \bbm \cos\psi &-\sin\psi & 0\\ \sin\psi & \cos\psi & 0\\ 0 & 0 & 1 \ebm, \\
	\mbf{R}_{xy}\mbf{R}_z\frac{\mbf{f}_{des}}{\norm{\mbf{f}}_{des}} &= \bbm 0 \\ 0 \\ 1 \ebm, \\
	\mbf{R}_{des} &= \mbf{R}_{xy} \mbf{R}_z	\\
	f_a &= \frac{1}{2}{\norm{\mbf{f}}_{des}}. 
\end{align}
The rotation error is then expressed as
\begin{align}
	\mbf{R}_{err} &= \mbf{R}_{est}(\mbf{R}_{des})^{-1}
\end{align}
The rotation error matrix $\mbf{R}_{err}$ is converted into Euler angles. $\Delta \phi$,
$\Delta\theta$, $\Delta\psi$, and desired body rates are derived based on a proportional control strategy,
\begin{equation}
	\boldsymbol{\omega}_{des} = \frac{1}{\mbf{\tau}_{att}}\times [\Delta\phi,\,\Delta\theta,\,\Delta\psi]^{T}.
\end{equation}

\subsubsection{Attitude Rate Control}

The rate controller first determines the rate error vector according to $
	\boldsymbol{\omega}_{err}=\boldsymbol{\omega}_{des} - \boldsymbol{\omega}_{est}$. The desired torque to apply to the vehicle is then obtained via a proportional-integral (PI) controller with a cross-coupling compensation term that results from the rigid-body dynamics,
\begin{equation}
	\begin{split}
	\mbf{\boldsymbol{m}}_{des}&=[\boldsymbol{m}_x, \boldsymbol{m}_y, \boldsymbol{m}_z]^{\rm T}\\
	&=\boldsymbol{\omega}_{est}\times\boldsymbol{J}\boldsymbol{\omega}_{est}+\frac{1}{\tau_{\omega}}\boldsymbol{J}\boldsymbol{\omega}_{err}+\mbf{K}_{I,\omega}\boldsymbol{J}\int\boldsymbol{\omega}_{err} \end{split},
\end{equation}
where $\boldsymbol{J}$ is the moment of inertia and $\boldsymbol{m}_{des}$ is the desired torque acting on the vehicle. The integral gain term is added to compensate for model bias and to reject disturbances. We have found that properly tuning the integral gain leads to much improved tracking results.

\subsubsection{Model Inverse} Given the desired torque and thrust vectors, we can solve for the corresponding actuator commands through feedback linearization, according to the nonlinear dynamic model given by \autoref{eq:dynamics}. Note that, to arrive at a simplified expression, we have ignored the drag terms, since they are small compared to propeller thrust in the hovering regime. The actuator commands are derived as follows:
\begin{align}
	\boldsymbol{\omega}_{left} &= \sqrt{\frac{\boldsymbol{m}_x + 2f_a l}{2k_tl}}, \\
	\boldsymbol{\omega}_{right} &= \sqrt{\frac{-\boldsymbol{m}_x + 2f_a l}{2k_tl}}, \\
	\delta_{left} &= \frac{-k_l k_t \boldsymbol{m}_y l^2 - k_p k_t \boldsymbol{m}_z l + k_m k_p k_t \boldsymbol{m}_x}{k_l k_p l (\boldsymbol{m}_x + 2 f_a l)}, \\
	\delta_{right} &= \frac{k_l k_t \boldsymbol{m}_y l^2 - k_p k_t \boldsymbol{m}_z l + k_m k_p k_t \boldsymbol{m}_x}{k_l k_p l (\boldsymbol{m}_x - 2 f_a l)}.
\end{align}

\subsubsection{Actuator Control}

Servo motors have built-in rotation feedback control, while, in general, the rotation rate of a brushless DC motor cannot be tracked directly. Instead, rate tracking is normally achieved by calibrating the relationship between motor voltage and rotation speed. However, this relationship varies with different motors, motor temperature, etc. We enable RPM tracking by allowing the ESCs to send pulses to the flight computer at phase commutations. We implement a PI controller to regulate the voltage delivered to each motor to eliminate any constant offset due to calibration errors. This enhances the motor response time to 25 ms, which results in greater bandwidth for high-level control.

\section{System Performance}
\label{sec:performance}

In order to supply practitioners with a benchmark for the expected performance of the Phoenix, we carried out a range of experiments in our Vicon motion capture facility. All controllers described in Section \ref{sec:control-arch} were running on-board the 168MHz Cortex M4F microprocessor for each experiment. The physical and controller parameters are listed in \autoref{table:params}. We used the existing, native PX4 implementation of a complementary filter-based attitude estimator for pose estimation. IMU measurements were low-pass filtered at a cut-off frequency of 20 Hz to obtain useful readings for the filter update/prediction stage. All high-level commands (for takeoff, landing, etc.) and motion capture updates were sent from our ground station running MAVROS to the drone using a wifi radio.

\subsection{Hovering}

\begin{figure}[h!]
	\includegraphics[width=\columnwidth]{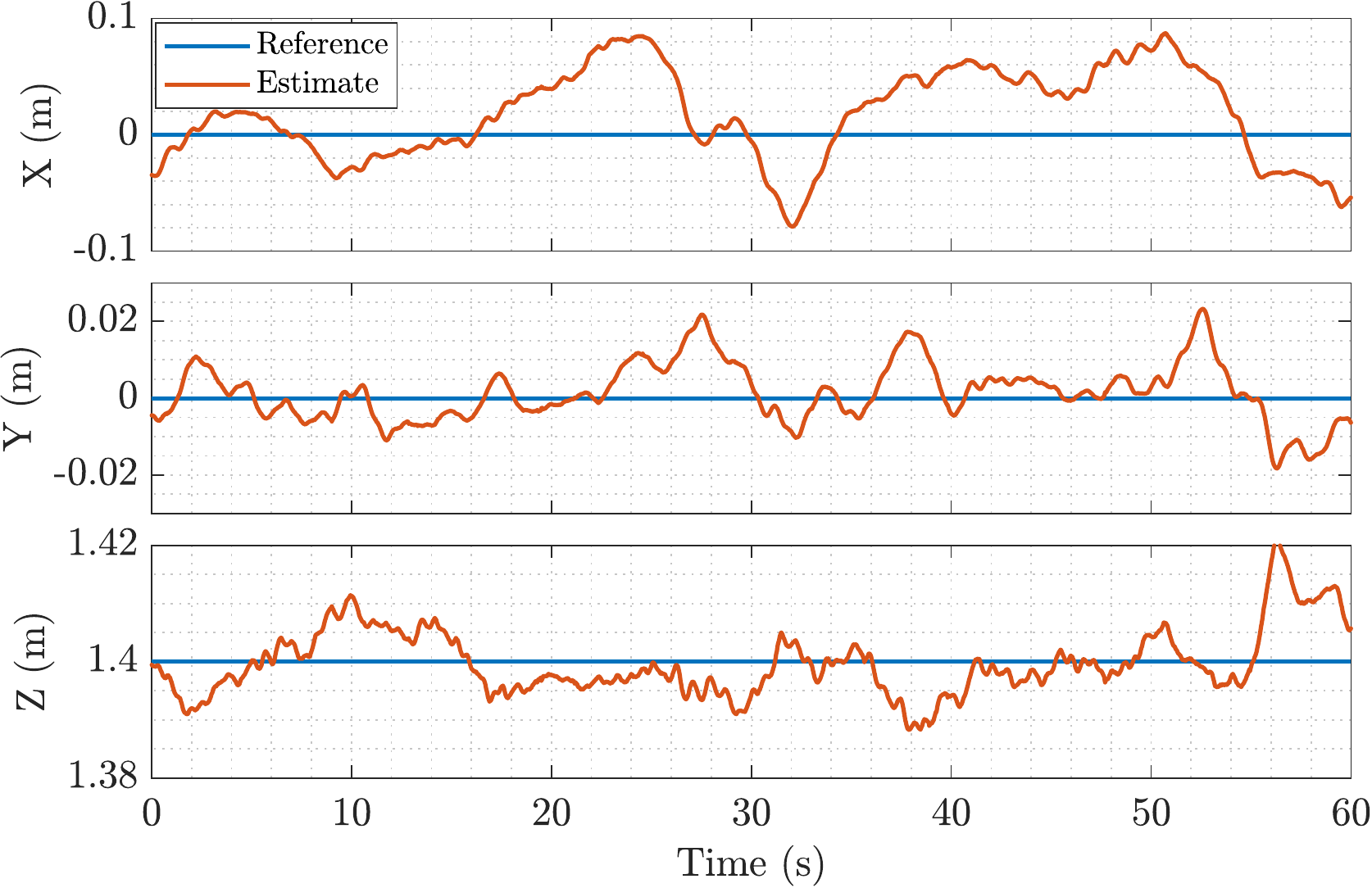}
	\caption{\small Vehicle $x$, $y$, and $z$ position during the one-minute fixed hovering experiment.}
	\label{fig:hovering-plot}
\end{figure}

We tested the hovering accuracy of the vehicle by commanding it to fly to a fixed position in the motion capture space and to hover in place for one minute. The desired position reference values and the estimates during hovering are plotted in \autoref{fig:hovering-plot}.

The vehicle is oriented such that the $\mbf{X}_{b}$ and $\mbf{Y}_{b}$ body frame axes align with the $\mbf{X}_{i}$ and $\mbf{Y}_{i}$ axes of the Vicon motion capture (inertial) frame. The RMS position errors in $x$, $y$, and $z$ were 4.3 cm, 0.8 cm, and 0.5 cm, respectively. Deviations along the body frame axis $\mbf{X}_{b}$ are the largest, due to the fact that the $x$ position of the vehicle is regulated by adjusting the thrust direction to match the desired acceleration direction. However, actuation along the pitch axis cannot be controlled as precisely as along the body roll axis, since there is larger aerodynamic uncertainty when actuating the control surfaces compared to altering the propeller speeds.

\subsection{Waypoint Transitions}

\begin{figure}[h!]
\begin{minipage}{0.225\textwidth}
\centering
	\includegraphics[width=1\linewidth]{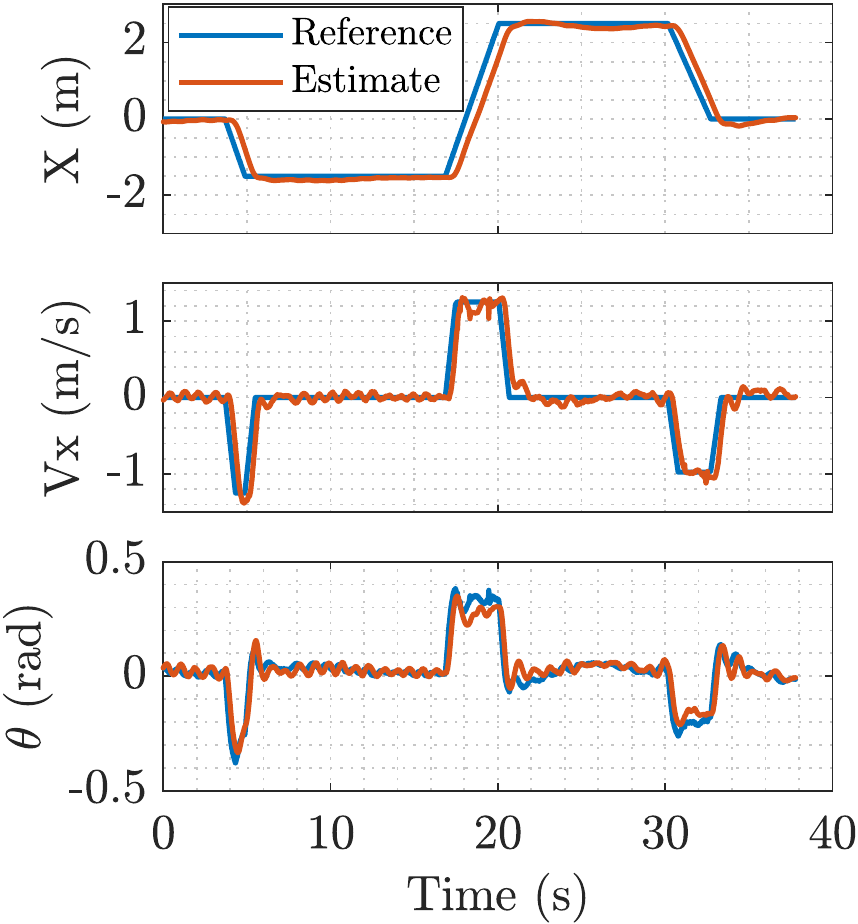}
	\caption{\small Pitch-based waypoint transition performance.}
	\label{fig:pitch-wp-plot}
\end{minipage}
\hfill
\begin{minipage}{.225\textwidth}
\centering
	\includegraphics[width=1\linewidth]{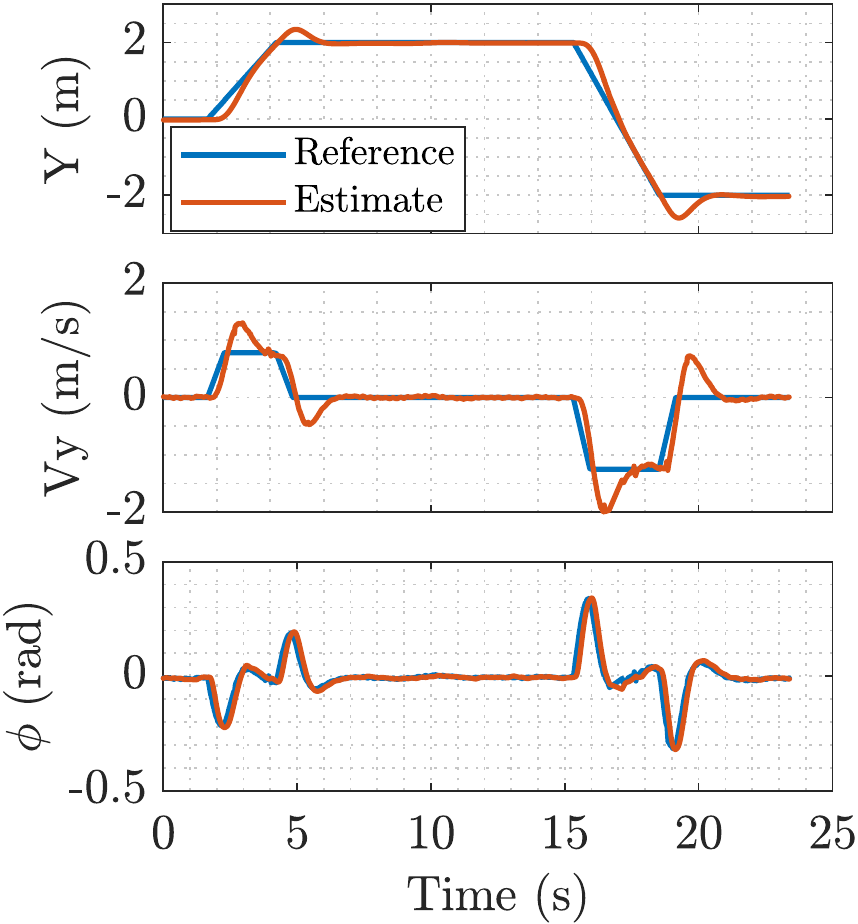}
	\caption{\small Roll-based waypoint transition performance.}
	\label{fig:roll-wp-plot}
\end{minipage}
\vspace{0.5\baselineskip}
\end{figure}

In \autoref{fig:pitch-wp-plot}, we visualize the position, velocity, and pitch angle tracking performance of the vehicle while performing a waypoint transition maneuver. Given a maximum commanded speed of 1.25 m/s, the vehicle pitches up to $17^{\circ}$ to maintain this speed during transition. In \autoref{fig:roll-wp-plot}, similar performance is demonstrated for a different waypoint transition involving body roll. 

\subsection{Trajectory Tracking}

\begin{figure}[h!]
\begin{minipage}{.20\textwidth}
\centering
	\vspace{-2mm}
	\includegraphics[width=\linewidth]{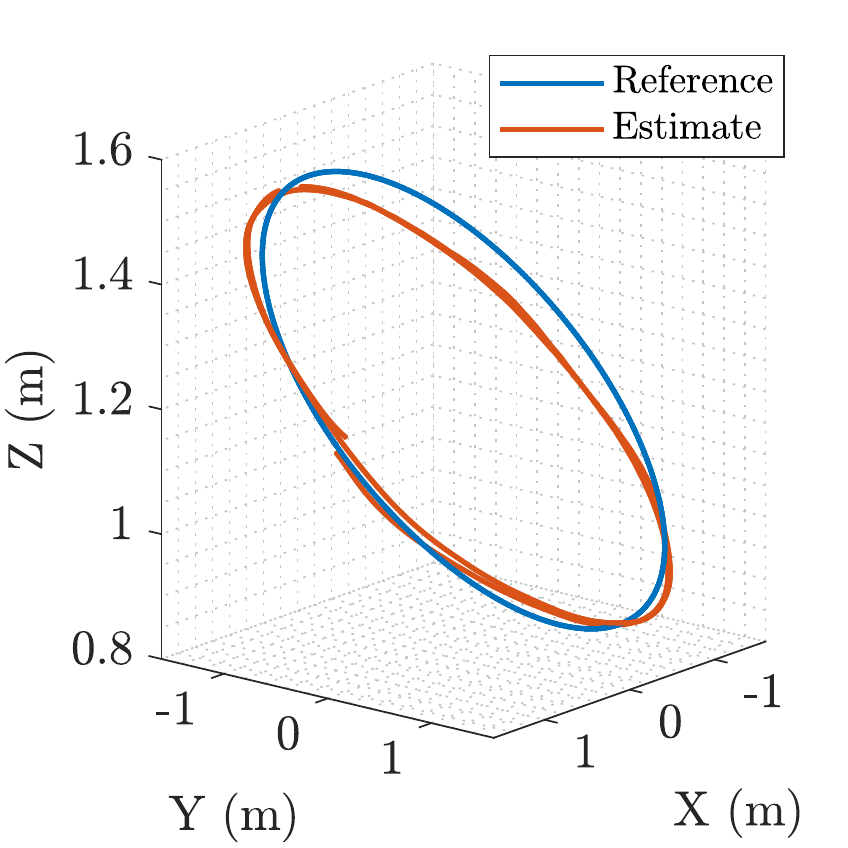}
	\caption{Circular trajectory tracking.}
	\label{fig:traj-3D-plot}
	\vspace{4mm}
	\includegraphics[width=0.90\linewidth]{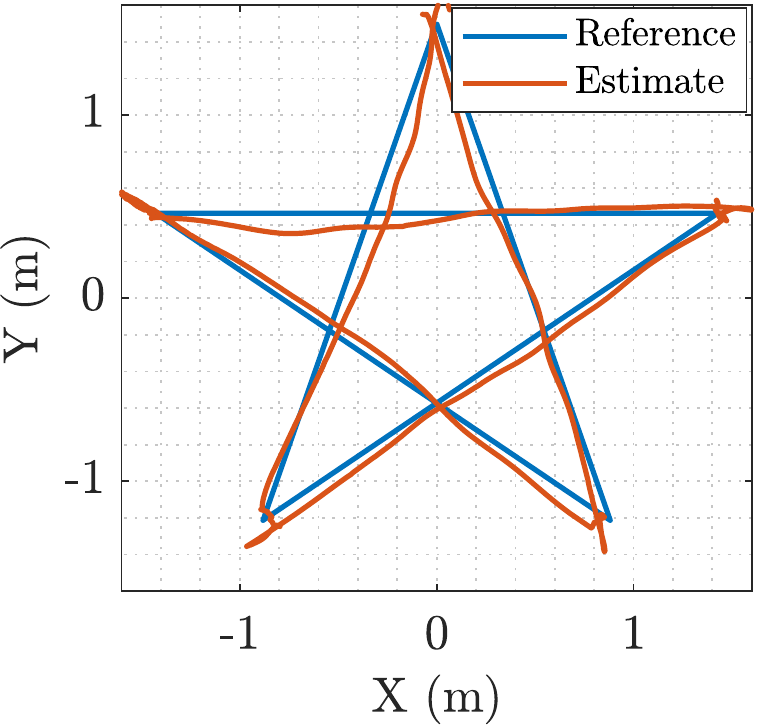}
	\caption{Horizontal star-shaped pattern trajectory tracking.}
	\label{fig:traj-STAR-plot}
\end{minipage}
\hfill
\begin{minipage}{.25\textwidth}
\centering
	\includegraphics[width=1\linewidth]{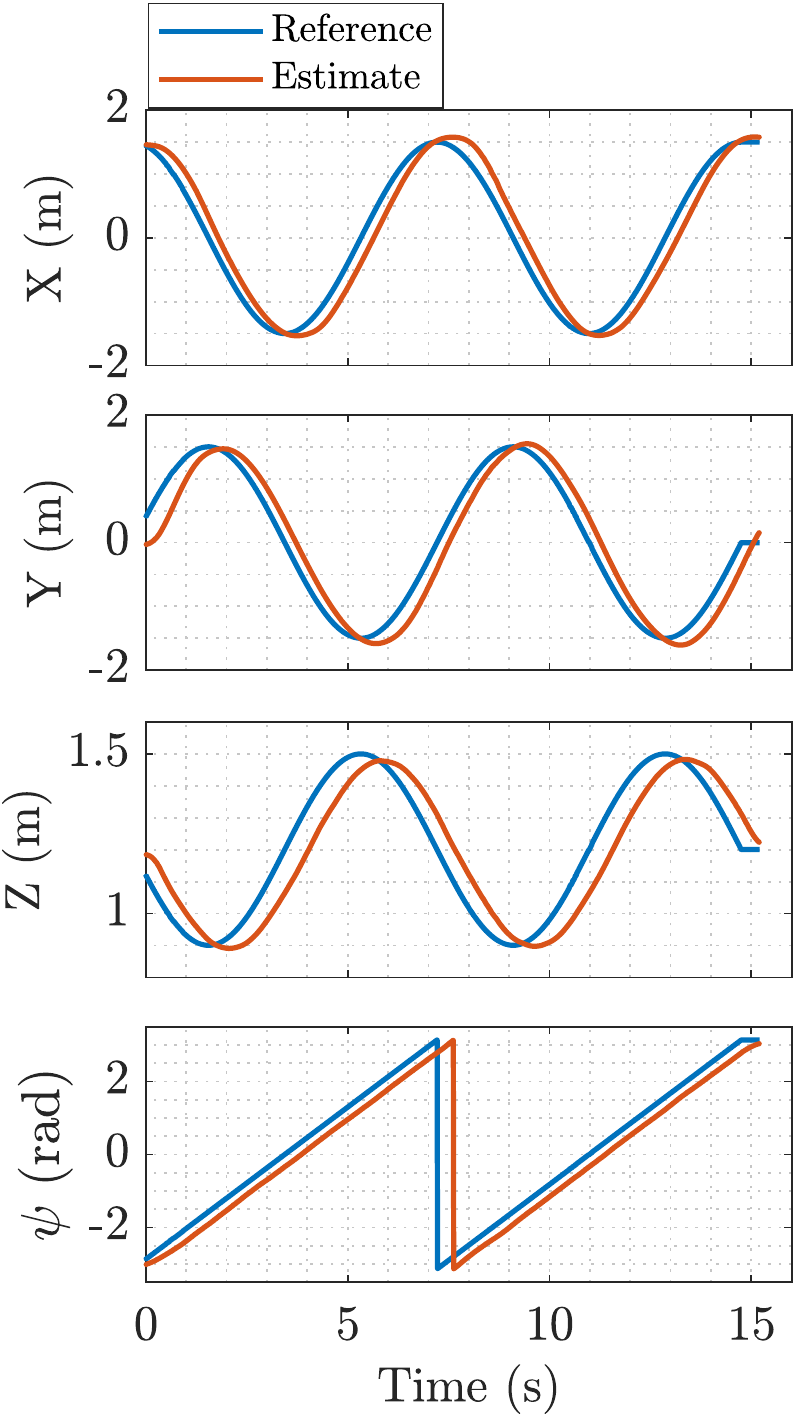}
	\caption{Vehicle $x$, $y$, $z$, and $\psi$ tracking during flight along circular trajectory.}
	\label{fig:traj-xyzy-plot}
\end{minipage}
\vspace{-\baselineskip}
\end{figure}

Finally, we evaluated the trajectory tracking performance of the vehicle by executing a circular trajectory in 3D with a radius of 1.5 m at a tracking speed of 1.5 m/s. Throughout the trajectory, pitch, roll, yaw, and thrust are all adjusted dynamically, verifying the agility of the platform. The 3D trajectory is plotted in \autoref{fig:traj-3D-plot}, and the corresponding $x$, $y$, $z$, and $\psi$ (yaw) estimates are shown in \autoref{fig:traj-xyzy-plot}. The reference signals are tracked well with an expected latency that results from the controller frequency response. In addition, \autoref{fig:traj-STAR-plot} shows the tracking results for a star-shaped trajectory in the horizontal plane.

\section{Open-Source Resources}
\label{sec:resources}

In this section, we briefly summarize the resources available in the open-source Phoenix drone repository. All of the code and the design documents can be found in our public GitHub repository located at \url{https://github.com/utiasSTARS/PhoenixDrone}. The overview below is not intended to be a comprehensive listing---more information is provided in the various repository \texttt{README.md} files.

\subsection{Mechanical CAD Files}

A single STEP file is provided that describes the overall mechanical design of the vehicle. This file can be imported into a wide variety of CAD tools. Additional STL files are also included for each of the 3D-printed components of the platform; end-users should be able to print the required parts and assemble the complete vehicle without requiring substantial CAD experience. Finally, a Bill of Materials sheet is available, which lists possible sources for all of the off-the-shelf components used to construct the drone.

\subsection{Autopilot Firmware}

The customized PX4 firmware that runs on the PixRacer autopilot is included in our release. The existing PX4 wiki incorporates a set of instructions to compile and upload the firmware to the PixRacer. Our firmware should also be compatible with other PX4-supported hardware with minor modifications to the \texttt{CMake} build files.

\subsection{ESC Firmware}

We have also released our modified ESC firmware, based on the BLHeli firmware, for use with the PX4 autopilot. This modified version allows the ESCs to send synchronized motor phase commutation signals to the autopilot, in order to monitor and regulate motor speeds precisely.

\subsection{Gazebo SITL Plugin}

\begin{figure}[h!]
	\vspace{3mm}
	\centering
	\includegraphics[width=\linewidth]{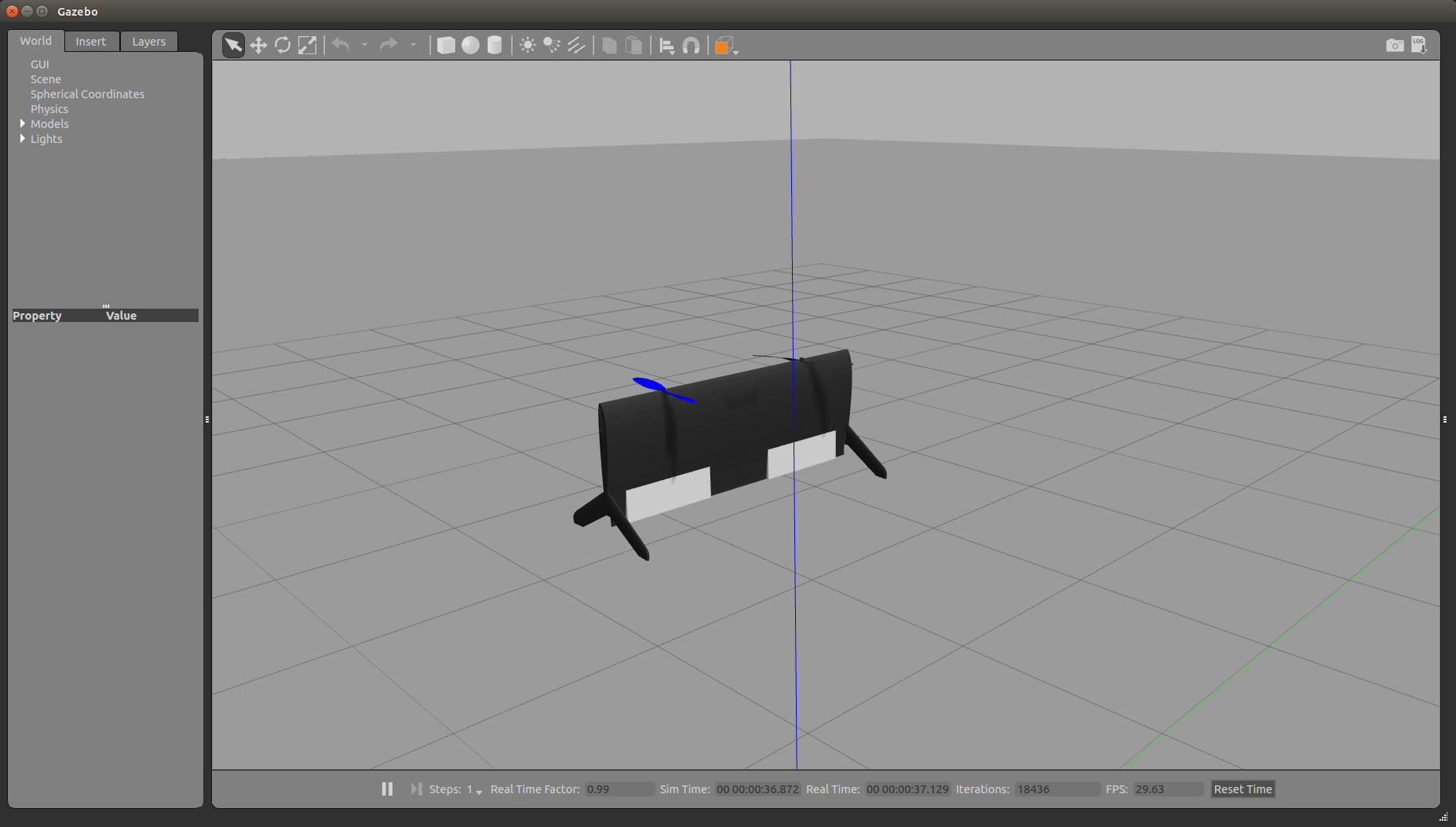}
	\caption{\small Example of the Gazebo client GUI during a software-in-the-loop simulation of the Phoenix vehicle in flight.}
	\label{fig:gazebo}
\vspace{-0.5\baselineskip}
\end{figure}

During development of the Phoenix, we found that STIL simulation was a crucial debugging tool. We have included the ROS Gazebo plug-ins required for SITL (software-in-the-loop) simulation using the PX4 firmware as part of our open-source package. SITL simulation allows end-users to verify their code behaviour in a virtual environment with realistic dynamics. Our customized Gazebo plug-ins simulate the dynamics of the vehicle based on a physical model extracted from real experimental data. An example view of the Gazebo GUI with the drone flying in simulation is shown in \autoref{fig:gazebo}.

\subsection{MATLAB Simulink Model}

We are also releasing the MATLAB Simulink model that captures the same controller architecture and high-fidelity dynamics of the vehicle. Compared to the SITL simulator, the Simulink environment is more flexible and allows for fast prototyping and verification of control laws, in addition to the probing of internal dynamic and control signals that are not easily accessible within the SITL simulation. 

\section{Use Cases and Future Work}
\label{sec:conclusion}

In this paper, we described the Phoenix drone, an open-source dual-rotor tail-sitter MAV designed for research and education. We provided a review of the system design (mechanics and modelling) including the controller architecture. Additionally, we presented a characterization of the performance of the vehicle in the hover and near-hover regimes. The unique characteristics of the platform make it suitable for both precision indoor trajectory tracking and for outdoor horizontal flight.

All of the associated design documents, schematics, and code have been released on GitHub under the permissive MIT licence. Our hope is that this release will encourage the aerial robotics community (researchers, educators, and hobbyists) to experiment, and to create innovative new modifications and derivatives. There are a number of possible use cases, ranging from preliminary research studies (in academia or industry) to teaching in a classroom environment (``Build Your Own Drone!''). To the best of our knowledge, there is no existing package that provides such a comprehensive and complete set of materials for tail-sitter development (royalty-free and completely open for use and modification). 

We are continuing to work with and extend the capabilities of the Phoenix. Presently, we are exploring new methods for vision-based mid-air docking and coordinated flight of two vehicles. We also plan to fully test and characterize the performance of the platform in forward flight, potentially with two vehicles in a docked configuration.

\section*{Acknowledgements}

The authors would like to thank the Aerospace Mechatronics Laboratory at McGill University for providing access to their load cell to conduct our dynamic modelling experiments.
\begin{table}[!h]
\centering
\renewcommand{\arraystretch}{1.32}
\caption{Parameter Table}
\begin{tabular}{p{1.2cm}|p{1.88cm}|p{4.4cm}}
\hline
\textbf{Parameter} & \textbf{Value (SI units)} & \textbf{Description}\\
\hline
$m$   & \num{0.65} & Vehicle mass \\
$l$   & \num{0.20} & Motor Arm Length to Centre of Mass \\
$b$   & \num{0.64} & Wing spaan \\
$\boldsymbol{J}_{xx}$ & \num{1.4e-2} & Body X axis Rotational Inertia\\
$\boldsymbol{J}_{yy}$ & \num{6.4e-3} & Body Y axis Rotational Inertia\\
$\boldsymbol{J}_{zz}$ & \num{1.8e-2} & Body Z axis Rotational Inertia\\
$k_t$ & \num{7.86e-6} & Propeller Thrust Constant\\
$k_m$ & \num{1.80e-7} & Propeller Torque Constant \\
$k_l$ & \num{3.48e-6} & Aerodynamic Lift Constant\\
$k_d$ & \num{1.75e-6} & Aerodynamic Drag Constant\\
$k_p$ & \num{3.44e-7} & Aerodynamic Pitch Moment Constant\\
$\tau_{p,xy}$ &  \num{0.5} & XY Position Controller Time Constant \\
$\tau_{p,z}$ & \num{0.3} & Z Position Controller Time Constant\\
$\zeta_{p,xy}$ & \num{0.6} & XY Position Controller Damping Ratio \\
$\zeta_{p,z}$ & \num{0.83} & Z Position Controller Damping Ratio \\
$\tau_{att}$ &\num{0.2} & Attitude Controller Time Constant \\
$\tau_{\boldsymbol{\omega},x}$ &\num{0.04} & Roll Rate Controller Time Constant \\
$\tau_{\boldsymbol{\omega},y}$ &\num{0.11} & Pitch Rate Controller Time Constant \\
$\tau_{\boldsymbol{\omega},z}$ &\num{0.04} & Yaw Rate Controller Time Constant \\
$\mbf{K}_{I,\omega,x}$ & \num{20.0} & Roll Rate Integral Gain\\
$\mbf{K}_{I,\omega,y}$ & \num{5.0} & Pitch Rate Integral Gain\\
$\mbf{K}_{I,\omega,z}$ & \num{0.0} & Yaw Rate Integral Gain\\
\end{tabular}
\label{table:params}
\end{table}

\bibliographystyle{IEEEcaps}

\IfFileExists{ts_paper.bib}
  {\bibliography{ts_paper.bib}} 
  {
} 

\end{document}